\documentclass[11pt]{article}

\usepackage{acl}[final]

\usepackage{times}
\usepackage{latexsym}

\usepackage[T1]{fontenc}

\usepackage[utf8]{inputenc}

\usepackage{microtype}

\usepackage{inconsolata}

\usepackage{graphicx}
\usepackage{amsmath}
\usepackage{amssymb}
\usepackage{booktabs}
\usepackage{multirow}
\usepackage{colortbl}
\usepackage{tabularx}
\definecolor{nestedkvblue}{RGB}{232,244,255}

\title{NestedKV: Nested Memory Routing for Long-Context KV Cache Compression}

\author{
\textbf{Hong Chen\textsuperscript{1}, Xiang Liu\textsuperscript{1}, Yubo Gao\textsuperscript{1}, Yuxuan Fan\textsuperscript{1}} \\
\textbf{Bo Wang\textsuperscript{1}, Yuanlin Chu\textsuperscript{1}, Yuanguo Lin\textsuperscript{2}, Xuming Hu\textsuperscript{1,*}} \\
\textsuperscript{1}The Hong Kong University of Science and Technology (Guangzhou), Guangzhou, China \\
\textsuperscript{2}Jimei University, Xiamen, China \\
\texttt{hchen763@connect.hkust-gz.edu.cn} \quad \texttt{xuminghu@hkust-gz.edu.cn}
}

\begin{document}
\maketitle

\begingroup
\renewcommand{\thefootnote}{}
\footnotetext{\textsuperscript{*}Corresponding author.}
\footnotetext{Code: \url{https://github.com/ChenHong30/NestedKV}}
\endgroup

\begin{abstract}
Long-context language models are limited by the memory footprint of the key-value (KV) cache. Existing training-free KV compression methods usually rank tokens by one importance signal --- attention, recency, layer-wise allocation, or key distinctiveness --- which becomes brittle when useful context is globally distinctive, locally episodic, or immediately relevant. We introduce NestedKV, a key-only KV cache compression method inspired by the Continuum Memory System in Nested Learning. NestedKV maintains global, block-level, and sliding-window key anchors, scores tokens by multi-time-scale cosine anomaly, and combines the resulting rankings with a training-free outer learner using head-adaptive mixing and surprise-gated token routing. The score is paired with adaptive per-head budgets and requires no training or LLM modification. Across RULER (4k--32k), LooGLE, LongBench, LongBench-E, InfiniteBench, and MMLU-Pro on Qwen3 and Llama-3.2 models, NestedKV is strongest when the retained cache is small. On Qwen3-4B, it improves over KeyDiff by up to 19.10 points on RULER and 19.29 on LongBench at $r=0.75$; at $r=0.95$, it retains 37.32 on LongBench versus 17.55 for KeyDiff.
\end{abstract}

\section{Introduction}

Long-context language models have become a standard interface for document understanding, retrieval-augmented generation, coding, and multi-turn interaction. Their practical deployment, however, is constrained by an increasingly simple bottleneck: the key-value (KV) cache grows linearly with context length and batch size. For long prompts and high-throughput serving, this transient memory can dominate inference cost even when model weights are fixed. As a result, a growing line of work studies training-free KV cache compression, aiming to reduce cache memory without fine-tuning the model or changing the attention implementation \citep{liu2023scissorhands,zhang2023h2o,xiao2024efficient,li2024snapkv,cai2024pyramidkv,feng2026ada,park2026keydiff}.

Most existing methods can be understood as choosing one anchor for token importance: past attention mass under a persistence-of-importance hypothesis \citep{liu2023scissorhands,zhang2023h2o}, recency and attention sinks \citep{xiao2024efficient}, an observation window near the end of the prompt \citep{li2024snapkv}, layer-wise cache budgets \citep{cai2024pyramidkv}, or geometric distinctiveness of keys from the mean direction \citep{park2026keydiff}. A complementary line allocates the cache budget adaptively across heads rather than uniformly \citep{feng2026ada}.

\begin{figure}[t]
  \centering
  \includegraphics[width=\columnwidth]{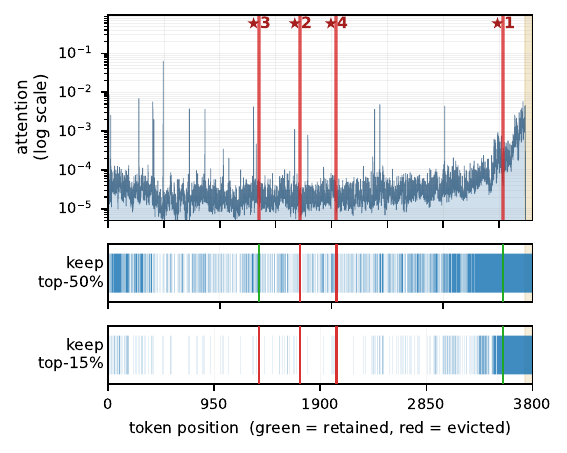}
  \caption{
    Attention from the last 64 queries on a long-context retrieval prompt (Qwen3-4B, RULER \texttt{niah\_multivalue}, $N{=}3{,}800$, 4 needles \mbox{$\star 1$--$\star 4$}). \textbf{Top}: attention mass (log scale). \textbf{Bottom}: tokens retained by an attention-sorted compressor at $r{=}0.50$ and $r{=}0.85$; surviving needles green, evicted red. This example is illustrative; Appendix~\ref{app:camera-ready} provides a dataset-level retention analysis.
  }
  \label{fig:intro_attention}
\end{figure}

Figure~\ref{fig:intro_attention} illustrates a possible mismatch between a past-attention proxy and the evidence that must remain available under aggressive compression. In this retrieval instance, attention concentrates on the prompt tail and the attention-sink prefix, while answer-bearing tokens receive little mass and can be evicted by an attention-sorted policy. The example motivates a complementary key-space signal; it is not by itself evidence that every attention-based compressor fails. Appendix~\ref{app:camera-ready} reports a dataset-level answer-span analysis at the same top-$k$ budget, which retains more retrieval evidence with NestedKV while also showing that answer-span retention alone does not determine every QA score.

These approaches are effective, but they also expose a common limitation: each method compresses the cache through a single view of memory. A token may be important because it is globally unusual in the document, because it marks a local topic shift inside one segment, or because it is part of the recent stream that will shape immediate generation. Under mild compression, a single statistic may be sufficient. Under aggressive compression or longer contexts, these notions diverge. A global mean can miss local episodes; a local rule can overfit repetitive blocks; a recent-window rule can discard earlier evidence needed for retrieval or multi-hop reasoning.

We propose NestedKV, a training-free KV cache compression method based on a continuum-memory view of token importance. Following the Nested Learning perspective that models maintain compressed context flows through nested memory systems with a self-modifying update rule \citep{behrouz2026nested}, NestedKV maintains a three-time-scale memory directly over the cached key stream --- a stable, an episodic, and a current anchor --- and scores each token by its cosine anomaly against each scale. The three scales act as inner learners: a token receives three rankings rather than one, and is retained if it is anomalous against any of the scales.

A training-free outer learner then combines these inner rankings on two axes. Per attention head, the most discriminative scale is up-weighted relative to a fixed prior, so heads can specialize in different temporal roles. Per token, the cross-scale disagreement between the three rankings is read as a compression-induced surprise signal, and high surprise smoothly routes the score from the blended view toward the strongest individual memory. Together, these two axes instantiate the self-modifying compressor motif of Nested Learning at test time, with no trainable parameters and no modification to the underlying LLM.

The score is key-only and remains compatible with optimized attention kernels. The full policy is combined with adaptive per-head memory allocation, separating two questions that are often entangled: which tokens are informative within a head, and how much memory each head should receive.

We evaluate NestedKV on a suite of long-context benchmarks (RULER \citep{hsieh2024ruler}, LongBench \citep{bai2024longbench}, LooGLE \citep{li2024loogle}, LongBench-E, and InfiniteBench \citep{zhang2024infty}) and a short-context knowledge benchmark (MMLU-Pro \citep{wang2024mmlu}), using Qwen3-4B as the primary frozen model. The main empirical pattern is that the continuum-memory score is most useful exactly where a single anchor should be weakest: at higher compression ratios and longer contexts. NestedKV is best or near-best across most RULER context–ratio cells, with the clearest gains under aggressive compression and longer contexts. It also improves the LongBench average from 30.77 to 50.06 at $r=0.75$, and on MMLU-Pro remains within $0.2$ points of the Full KV ceiling at $r{=}0.25$ while most baselines degrade.

Our contributions are:
\begin{itemize}
    \item We reframe training-free KV cache compression as continuum-memory anomaly detection over the key stream, giving token eviction a Nested Learning interpretation as bounded test-time memory maintenance.
    \item We introduce NestedKV, which uses three time-scale key statistics --- stable, episodic, and current --- as inner learners and combines their per-token anomaly rankings through a training-free outer learner that adapts per head and per token, the latter driven by compression-induced surprise. The score is paired with adaptive per-head memory allocation, with no training or LLM modification.
    \item We provide empirical evidence across six benchmarks --- RULER, LongBench, LooGLE, LongBench-E, InfiniteBench, and MMLU-Pro --- that multi-time-scale scoring is especially valuable under aggressive compression and long contexts, while not compromising short-prompt capability.
\end{itemize}

\section{Method}

\begin{figure*}[t]
  \centering
  \includegraphics[width=\textwidth]{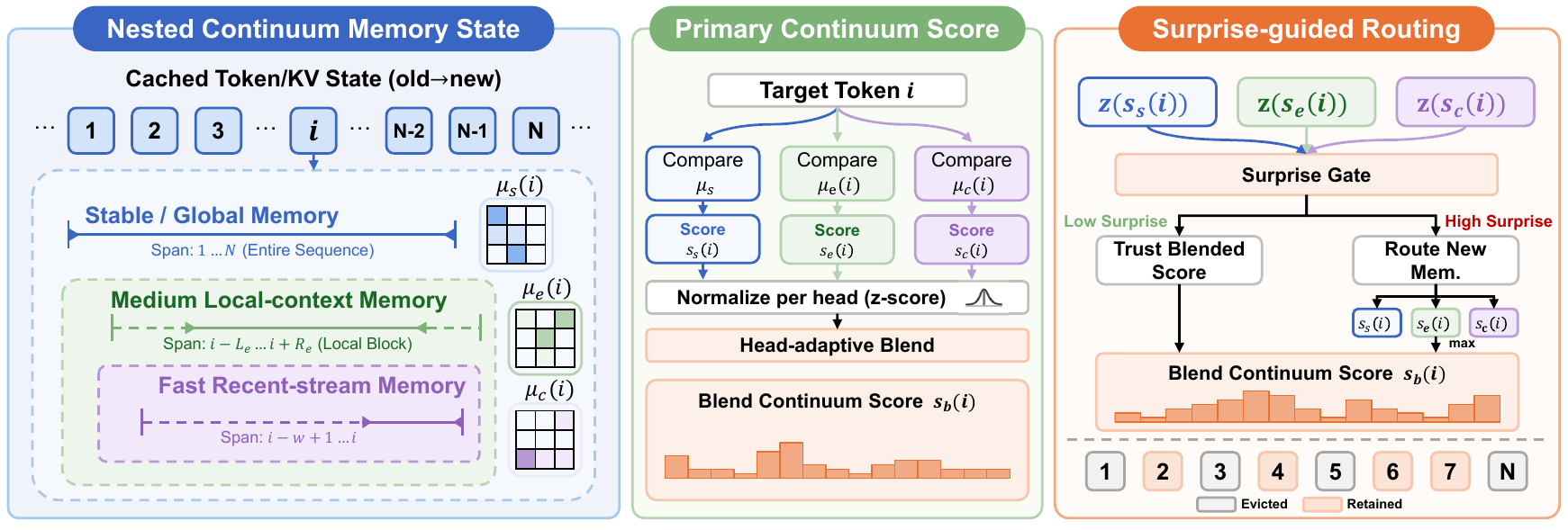}
  \caption{
    Overview of NestedKV.
    \textbf{Left (Section~\ref{sec:memory}).} Three time-scale summaries of the cached key stream: stable mean $\mu_s$, episodic block mean $\mu_e(i)$, and current sliding-window mean $\mu_c(i)$.
    \textbf{Middle (Sections~\ref{sec:redundancy}--\ref{sec:surprise}).} Each key produces per-scale cosine anomalies $s_s(i), s_e(i), s_c(i)$, normalized per head and combined by a head-adaptive softmax into the blended score $s_b(i)$.
    \textbf{Right (Section~\ref{sec:surprise}).} Surprise-guided routing measures cross-scale disagreement: agreeing tokens keep $s_b(i)$, disagreeing tokens are routed to the maximum-anchor reading so any single anomaly flag suffices. The final score drives the retain/evict decision (bottom row).
  }
  \label{fig:framework}
\end{figure*}

NestedKV compresses the KV cache after prefill and before autoregressive decoding. It is applied independently at each Transformer layer, while coordinating memory allocation across the layer's KV heads. The model parameters, attention function, and retained value vectors are unchanged; the method only determines which cached positions remain in the bounded test-time memory. Figure~\ref{fig:framework} summarizes the three components: a nested continuum memory state over the cached keys, a per-scale anomaly score blended into a primary continuum reading, and a surprise-guided routing rule that selects between the blended reading and the strongest individual memory for each token.

\subsection{KV Compression as Nested Memory Maintenance}

For a frozen LLM, the prefilled KV cache is the inner memory state through which the model carries the context flow into future decoding steps. KV compression therefore asks for a bounded memory policy rather than a standalone token deletion rule. For layer $\ell$ and KV head $h$, let
\begin{equation}
M_{\ell,h} = (K_{\ell,h}, V_{\ell,h})
\end{equation}
be the full prefill memory. NestedKV constructs a compressed memory
\begin{equation}
M_{\ell,h}^{B_h}
= \mathcal{C}_{\phi}(K_{\ell,h},V_{\ell,h};B_h),
\end{equation}
where $B_h$ is the head-specific memory budget and $\phi$ denotes the fixed NestedKV memory policy. No parameters in $\phi$ are learned; the policy is defined by the continuum-memory state and the head-wise allocation rule below.

To simplify notation, we describe one layer and one KV head and omit $\ell,h$. Let
\begin{equation}
\begin{aligned}
K &= [k_1,\ldots,k_N] \in \mathbb{R}^{N \times d},\\
V &= [v_1,\ldots,v_N] \in \mathbb{R}^{N \times d_v}.
\end{aligned}
\end{equation}
Given budget $B$, the compressor returns an index set $\mathcal{S}$ with $|\mathcal{S}|=B$ and memory $M^B=(K_{\mathcal{S}},V_{\mathcal{S}})$. All scores are computed from normalized keys $\hat{k}_i = k_i/\lVert k_i\rVert_2$, so the memory policy focuses on directional structure in key space.

\begin{figure}[t]
  \centering
  \includegraphics[width=0.97\columnwidth]{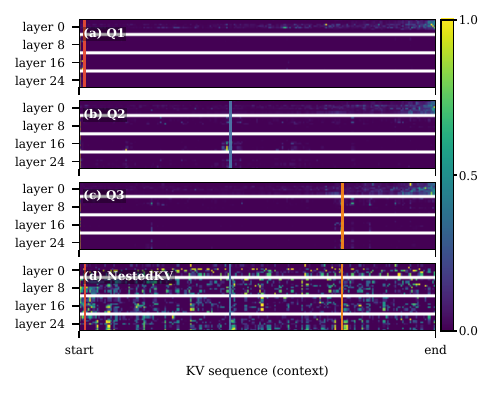}
  \caption{LongBench-Qasper attention-series probe \citep{dasigi2021dataset,bai2024longbench}. Q1--Q3 attend to different answer regions (vertical lines), while NestedKV assigns saliency across these dispersed positions.}
  \label{fig:method-attention-series}
\end{figure}

Figure~\ref{fig:method-attention-series} shows the motivation for using a memory state rather than a single token-importance view. Even within the same Qasper document, different downstream questions activate different answer regions across layers. A compressor that commits to one temporal anchor risks preserving only one such pattern. NestedKV therefore constructs a continuum memory over the key stream, so a token can be retained when it is distinctive globally, within its local episode, or relative to the current stream.

\subsection{Continuum Memory State}
\label{sec:memory}

The Continuum Memory System view suggests that memory should not collapse into a single temporal scale. A token can be redundant with the document as a whole, redundant within its local episode, or redundant with the recent stream. NestedKV represents these three notions as a continuum memory state
\begin{equation}
\mathcal{M}(i)=\{\mu_s,\mu_e(i),\mu_c(i)\}
\end{equation}
for every cached token $i$.

\paragraph{Stable memory.}
The stable component summarizes the whole prefilled context:
\begin{equation}
\mu_s = \frac{1}{N}\sum_{j=1}^{N} \hat{k}_j .
\end{equation}
It captures document-level regularities that persist across the entire context.

\paragraph{Episodic memory.}
The episodic component summarizes the local segment containing token $i$. Let $B(i)$ be the block containing $i$, with block size
\begin{equation}
b = \operatorname{clip}\left(\left\lfloor N/32 \right\rfloor, 128, 256\right).
\end{equation}
Then
\begin{equation}
\mu_e(i) = \frac{1}{|B(i)|}\sum_{j \in B(i)} \hat{k}_j .
\end{equation}
It captures passage-level or turn-level structure that may be invisible to a global summary.

\paragraph{Current memory.}
The current component summarizes the immediate causal stream ending at $i$. With window size $W=64$,
\begin{equation}
\begin{aligned}
\mu_c(i) &=
\frac{1}{i-\ell_i+1}\sum_{j=\ell_i}^{i} \hat{k}_j,\\
\ell_i &= \max(1, i-W+1).
\end{aligned}
\end{equation}
It captures short-range continuity and immediate redundancy.

\subsection{Per-Scale Anomaly Scores}
\label{sec:redundancy}

NestedKV evicts a token when its key is already predictable from the continuum memory state and retains it when it is anomalous under that state. Because the three memory scales summarize different temporal structure, we keep their anomaly readings separate rather than collapsing them into a single anchor up front. For each cached token $i$, the per-scale anomaly scores are
\begin{equation}
\begin{aligned}
a_s(i) &= -\cos(\hat{k}_i,\mu_s),\\
a_e(i) &= -\cos(\hat{k}_i,\mu_e(i)),\\
a_c(i) &= -\cos(\hat{k}_i,\mu_c(i)).
\end{aligned}
\end{equation}
A low $a_k(i)$ means token $i$ is typical with respect to memory scale $k$; a high $a_k(i)$ means the token carries information not well explained by that scale and should remain available for future attention. Each score is min-max normalized within its head, yielding $\tilde{a}_s,\tilde{a}_e,\tilde{a}_c$ on a common per-head scale. To preserve attention stability, the first $n_{\mathrm{sink}}=4$ positions are pinned by assigning them a large value before selection.

The three readings now propose three potentially different token rankings, and any rule for combining them is itself a modeling decision. The next subsection introduces the outer learner that performs this combination.

\subsection{Outer Learner: Head-Adaptive Blend with Surprise Routing}
\label{sec:surprise}

NestedKV combines the per-scale anomaly scores through a training-free outer learner that adapts on two complementary axes: which memory scale is reliable \emph{on each attention head}, and which combination rule should apply \emph{on each token}.

\paragraph{Head-adaptive blend.}
Different heads specialize in different temporal roles, so a single mixing rule across all heads wastes head capacity. For each head we measure how strongly each memory scale separates its top from its bottom tokens,
\begin{equation}
\Delta_k = \overline{\operatorname{top}_p(\tilde{a}_k)} - \overline{\operatorname{bot}_p(\tilde{a}_k)},
\end{equation}
with $p=10\%$. A larger $\Delta_k$ means scale $k$ produces a more discriminative ranking on this head. The head-adaptive blend weight is then a softmax over reliability gaps anchored by a fixed log-prior,
\begin{equation}
w_k = \frac{\exp\bigl(\log w_k^0 + \beta\,\Delta_k\bigr)}{\sum_{j}\exp\bigl(\log w_j^0 + \beta\,\Delta_j\bigr)},
\end{equation}
with prior $(w_s^0, w_e^0, w_c^0) = (0.4, 0.4, 0.2)$ shared across the model and fixed temperature $\beta$. The blended score on that head is
\begin{equation}
a_{\text{blend}}(i) = w_s\,\tilde{a}_s(i) + w_e\,\tilde{a}_e(i) + w_c\,\tilde{a}_c(i).
\end{equation}
The prior $(0.4, 0.4, 0.2)$ enters only as a Bayesian anchor, not as a final coefficient: heads on which one scale is markedly more discriminative shift weight onto that scale, while heads on which the three scales are similarly informative fall back on the prior.

\paragraph{Compression-induced surprise.}
The blended score is reliable when the three memory scales agree on the relative anomaly of a token, but it is brittle when they disagree. A token that is highly anomalous against the stable memory may still be typical inside its local episode, or vice versa, and any average hides exactly the cross-scale information that distinguishes these cases. We define the \emph{compression-induced surprise} of token $i$ as the standard deviation of the inner memories' rankings,
\begin{equation}
s(i) = \operatorname{std}\bigl(\tilde{a}_s(i),\,\tilde{a}_e(i),\,\tilde{a}_c(i)\bigr).
\end{equation}
Low surprise means the inner memories agree; high surprise means they disagree, so any single average is at risk of being driven by one scale's blind spot.

\paragraph{Routed score.}
When the inner memories disagree, the safer reading is the strongest individual memory rather than their average:
\begin{equation}
a_{\text{win}}(i) = \max\bigl(\tilde{a}_s(i),\,\tilde{a}_e(i),\,\tilde{a}_c(i)\bigr).
\end{equation}
The NestedKV score combines the two branches with a sigmoid gate over surprise,
\begin{equation}
\begin{aligned}
\alpha(i) &= \sigma\bigl(\kappa\,(s(i)-\tau)\bigr),\\
a^{\star}(i) &= (1-\alpha(i))\,a_{\text{blend}}(i) + \alpha(i)\,a_{\text{win}}(i),
\end{aligned}
\end{equation}
with fixed gate threshold $\tau$ and sharpness $\kappa$ shared across all benchmarks. Tokens on which the three scales agree pass through the head-adaptive blend; tokens that produce high cross-scale disagreement are routed toward the strongest memory.


\subsection{NestedKV Compression Operator}

The single-head compression operator applies $\operatorname{TopB}$ to the NestedKV score $a^{\star}$:
\begin{equation}
\mathcal{C}_{\phi}(K,V;B)
= \{(k_i,v_i): i\in \operatorname{TopB}(a^{\star}_{1:N})\},
\end{equation}
where $\operatorname{TopB}$ returns the $B$ positions with the largest scores after sink pinning. This operator is the concrete bounded-memory update of NestedKV: it keeps positions that are anomalous against at least one memory scale --- either by the head-adaptive blend or by the surprise-routed winner --- and removes positions already absorbed by the continuum state.

\subsection{Head-Wise Memory Competition}

Different heads act as parallel memory channels and can specialize in different temporal roles. Some heads may concentrate high residuals around local episodes, while others may spread memory over stable document structure or recent transitions. NestedKV therefore allocates a layer budget across heads by letting head-token pairs compete under their normalized continuum residuals.

Let $a_{h,i}$ be the normalized residual for token $i$ in head $h$, and let $B_\ell$ be the total number of KV positions retained in layer $\ell$. NestedKV selects the globally highest-residual pairs
\begin{equation}
\mathcal{P}_{\ell}
= \operatorname{TopB}_{B_\ell}\{(h,i): a_{h,i}\},
\end{equation}
subject to a small per-head safeguard so that every memory channel keeps a minimum state. This induces a head-specific budget
\begin{equation}
B_h = |\{i:(h,i)\in \mathcal{P}_{\ell}\}|,
\quad \sum_h B_h=B_\ell.
\end{equation}
Each head then applies $\mathcal{C}_{\phi}(K_h,V_h;B_h)$. The result is a bounded layer memory whose capacity is distributed according to continuum-memory surprise rather than a uniform allocation.

\section{Experiments}
\label{sec:experiments}

\begin{table*}[t]
\centering
\scriptsize
\setlength{\tabcolsep}{1.8pt}
\resizebox{\textwidth}{!}{%
\begin{tabular}{llcccccccccccc}
\toprule
\multirow{2}{*}{Model} & \multirow{2}{*}{Method} & \multicolumn{3}{c}{4k} & \multicolumn{3}{c}{8k} & \multicolumn{3}{c}{16k} & \multicolumn{3}{c}{32k} \\
\cmidrule(lr){3-5}\cmidrule(lr){6-8}\cmidrule(lr){9-11}\cmidrule(lr){12-14}
& & $r=.25$ & $r=.50$ & $r=.75$ & $r=.25$ & $r=.50$ & $r=.75$ & $r=.25$ & $r=.50$ & $r=.75$ & $r=.25$ & $r=.50$ & $r=.75$ \\
\midrule
\multirow{7}{*}{Qwen3-4B} & Full KV & \multicolumn{3}{c}{93.83} & \multicolumn{3}{c}{91.65} & \multicolumn{3}{c}{91.13} & \multicolumn{3}{c}{85.78} \\
& PyramidKV & 80.61 & 34.45 & 25.72 & 68.37 & 41.83 & 23.85 & 77.80 & 44.92 & 29.76 & 69.15 & 50.47 & 36.82 \\
& ExpAttn & 79.48 & 50.26 & 22.64 & 78.28 & 53.77 & 26.34 & 74.73 & 45.91 & 30.44 & 85.01 & 82.52 & 66.98 \\
& SnapKV & 85.02 & 58.23 & 28.42 & 73.85 & 56.03 & 33.27 & 81.68 & 60.79 & 37.95 & 73.85& 62.33 & 47.14 \\
& StreamingLLM & 77.72 & 59.53 & 37.53 & 76.49 & 55.99 & 33.99 & 70.00 & 52.38 & 34.20 & 66.04 & 47.21 & 29.17 \\
& KeyDiff & 89.12 & 75.06 & 60.22 & 82.72 & 74.36 & 62.53 & 76.84 & 72.51 & 62.20 & 68.62 & 63.38 & 55.77 \\
\rowcolor{nestedkvblue} & NestedKV & \textbf{93.60} & \textbf{90.40} & \textbf{79.32} & \textbf{91.11} & \textbf{89.35} & \textbf{78.52} & \textbf{90.44} & \textbf{88.36} & \textbf{79.89} & \textbf{85.25} & \textbf{82.61} & \textbf{73.11} \\
\midrule
\multirow{7}{*}{Qwen3-8B} & Full KV & \multicolumn{3}{c}{94.52} & \multicolumn{3}{c}{93.71} & \multicolumn{3}{c}{93.07} & \multicolumn{3}{c}{89.56} \\
& PyramidKV & 75.37 & 40.48 & 29.49 & 69.62 & 42.95 & 27.76 & 77.05 & 49.75 & 35.85 & 74.85 & 54.63 & 42.76 \\
& ExpAttn & 75.37 & 52.64 & 22.18 & 79.89 & 54.16 & 23.14 & 76.60 & 50.20 & 30.81 & 88.15 & 82.58 & 75.54 \\
& SnapKV & 84.00 & 56.33 & 32.09 & 74.11 & 55.35 & 35.42 & 82.19 & 63.34 & 45.28 & 78.11 & 64.20 & 50.25 \\
& StreamingLLM & 79.11 & 60.97 & 39.95 & 78.71 & 59.41 & 38.12 & 73.57 & 54.55 & 35.81 & 70.89 & 50.97 & 31.65 \\
& KeyDiff & 88.89 & 77.21 & 65.53 & 85.17 & 77.65 & 67.74 & 84.65 & 76.54 & 66.49 & 75.00 & 70.32 & 62.42 \\
\rowcolor{nestedkvblue} & NestedKV & \textbf{94.73} & \textbf{93.24} & \textbf{80.66} & \textbf{94.24} & \textbf{91.26} & \textbf{80.81} & \textbf{91.46} & \textbf{88.10} & \textbf{82.67} & \textbf{88.75} & \textbf{85.41} & \textbf{76.17} \\
\midrule
\multirow{7}{*}{Llama-3.2-1B-Inst.} & Full KV & \multicolumn{3}{c}{75.50} & \multicolumn{3}{c}{70.14} & \multicolumn{3}{c}{69.19} & \multicolumn{3}{c}{66.90} \\
& PyramidKV & 60.35 & 21.48 & 20.67 & 47.50 & 22.53 & 14.90 & 46.90 & 22.48 & 14.50 & 50.23 & 36.09 & 27.99 \\
& ExpAttn & 61.48 & 42.90 & 19.15 & 55.75 & 35.15 & 18.52 & 49.47 & 31.25 & 16.80 & 63.68 & 58.98 & 38.18 \\
& SnapKV & 63.03 & 42.24 & 20.78 & 49.16 & 33.28 & 19.79 & 55.04 & 37.41 & 22.72 & 53.35 & 43.22 & 31.70 \\
& StreamingLLM & 61.34 & 42.80 & 24.76 & 58.00 & 40.60 & 21.99 & 52.15 & 37.47 & 24.23 & 50.50 & 35.61 & 20.59 \\
& KeyDiff & \textbf{72.39} & 66.26 & 56.55 & 67.55 & \textbf{61.88} & 49.55 & 65.26 & 61.73 & 49.74 & 64.41 & 58.93 & 47.96 \\
\rowcolor{nestedkvblue} & NestedKV & 72.28 & \textbf{68.74} & \textbf{60.53} & \textbf{67.71} & 61.42 & \textbf{51.88} & \textbf{66.99} & \textbf{62.48} & \textbf{54.20} & \textbf{64.40} & \textbf{60.29} & \textbf{50.93} \\
\midrule
\multirow{7}{*}{Llama-3.2-3B-Inst.} & Full KV & \multicolumn{3}{c}{91.80} & \multicolumn{3}{c}{83.99} & \multicolumn{3}{c}{84.04} & \multicolumn{3}{c}{77.26} \\
& PyramidKV & 73.16 & 43.13 & 28.19 & 62.74 & 45.47 & 26.22 & 65.22 & 46.82 & 29.51 & 60.27 & 47.87 & 34.81 \\
& ExpAttn & 80.59 & 65.50 & 40.18 & 72.77 & 58.71 & 38.00 & 73.00 & 60.00 & 39.64 & 74.88 & 49.88 & 30.63 \\
& SnapKV & 73.84 & 49.73 & 28.37 & 68.71 & 52.12 & 30.95 & 68.84 & 53.87 & 34.98 & 68.36 & 57.06 & 42.17 \\
& StreamingLLM & 77.78 & 56.70 & 33.09 & 70.16 & 50.01 & 28.42 & 66.63 & 50.56 & 27.23 & 61.64 & 43.40 & 25.70 \\
& KeyDiff & 88.69 & 82.32 & 70.09 & 81.20 & 74.75 & 67.29 & 79.30 & 73.47 & 66.46 & 75.32 & \textbf{71.91} & 62.55 \\
\rowcolor{nestedkvblue} & NestedKV & \textbf{90.16} & \textbf{85.48} & \textbf{76.90} & \textbf{83.86} & \textbf{80.57} & \textbf{70.88} & \textbf{81.05} & \textbf{76.83} & \textbf{69.13} & \textbf{75.40} & 71.28 & \textbf{62.60} \\
\bottomrule
\end{tabular}}
\caption{RULER average over 13 tasks. Each group reports Full KV and compressed-cache results across model scales, context lengths, and eviction ratios. Shaded rows are NestedKV.}
\label{tab:ruler-main}
\end{table*}

We evaluate NestedKV as a training-free KV cache compressor for frozen LLM inference. The experimental matrix tests whether continuum-memory compression is robust across model families, model scales, compression ratios, and task types.

\subsection{Experimental Setup}

\paragraph{Models.}
We evaluate Qwen3-0.6B, Qwen3-4B, Qwen3-8B \citep{yang2025qwen3}, Llama-3.2-1B-Instruct, and Llama-3.2-3B-Instruct \citep{grattafiori2024llama}. To test scaling beyond the main model suite, we additionally evaluate Qwen3-32B on the full 13-task RULER suite at 4k and 8k contexts. For every benchmark and compression ratio, we also run the same model with the full KV cache as an upper-bound reference.

\paragraph{Benchmarks.}
We evaluate on a suite of long-context and short-context benchmarks. RULER \citep{hsieh2024ruler} provides controlled synthetic tasks across multiple context lengths; we report the average over 13 RULER tasks at 4k, 8k, 16k, and 32k contexts. LongBench \citep{bai2024longbench} evaluates real long-context understanding tasks. LooGLE \citep{li2024loogle} evaluates long-dependency and short-dependency QA over long documents. We additionally evaluate LongBench-E \citep{bai2024longbench} and InfiniteBench longbook\_qa\_eng and code\_debug \citep{zhang2024infty}; to test whether the compressor preserves short-prompt capability, we evaluate MMLU-Pro \citep{wang2024mmlu}, a 10-choice multi-domain knowledge benchmark. The LongBench-E and InfiniteBench numbers are reported in Appendix~\ref{app:additional}.

\paragraph{Compression ratios.}
We evaluate eviction ratios $r \in \{0.25,0.50,0.75\}$, where $r$ denotes the fraction of KV entries removed after prefill. Unless otherwise stated, NestedKV uses one fixed block schedule, current-memory window, sink-token count, and routing configuration across models, benchmarks, and compression ratios; the corresponding sensitivity analyses are reported in Appendix~\ref{app:camera-ready}.

\paragraph{Baselines.}
We compare against representative training-free KV compression methods implemented through the \texttt{kvpress} evaluation framework released with Expected Attention \citep{devoto2025expected}: StreamingLLM \citep{xiao2024efficient}, SnapKV \citep{li2024snapkv}, Expected Attention, PyramidKV \citep{cai2024pyramidkv}, and KeyDiff \citep{park2026keydiff}. Full-cache inference is included as the no-compression reference. We do not include H$_2$O \citep{zhang2023h2o} in the main accuracy matrix because attention-history methods are memory-infeasible in much of our eager-attention evaluation matrix. Appendix~\ref{app:camera-ready} reports a separate ObservedAttention (H$_2$O-like) feasibility study, including OOM frequencies rather than treating failed runs as zero-valued scores.

\begin{figure*}[!t]
  \centering
  \includegraphics[width=\textwidth]{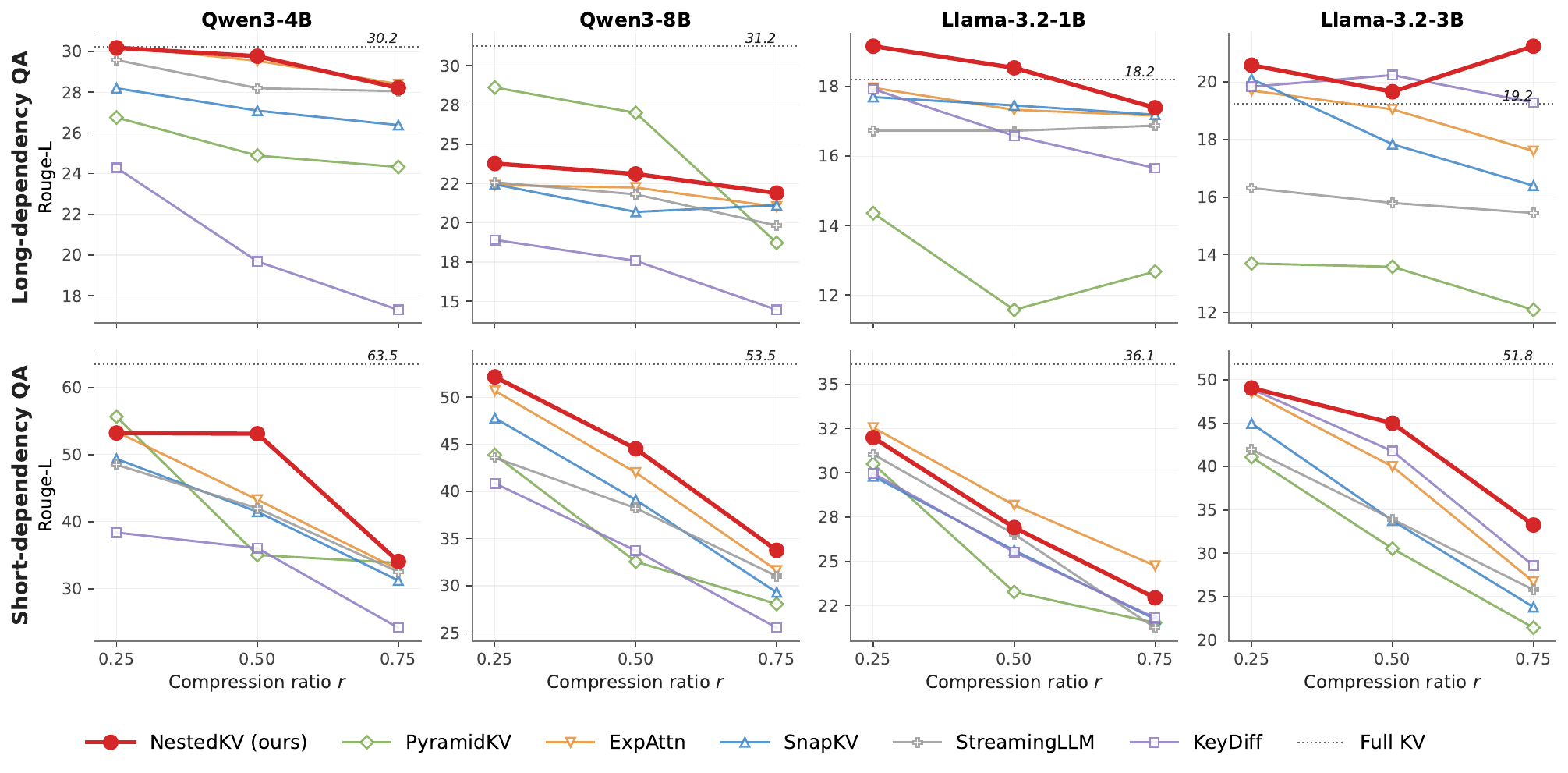}
  \caption{LooGLE Rouge-L score as a function of the eviction ratio $r$ for the long-dependency QA (top) and short-dependency QA (bottom) splits, across four models. Dotted lines mark each model's Full KV reference. NestedKV (solid red) is competitive across dependency regimes and remains robust under stronger compression.}
  \label{fig:loogle-lines}
\end{figure*}

\paragraph{Environment.}
All experiments are conducted on a server with four NVIDIA L20 GPUs. We use the same hardware configuration for full-cache references and compressed-cache runs so that accuracy comparisons are not affected by changes in model parallelism or execution backend.

\begin{figure}[t]
  \centering
  \includegraphics[width=\columnwidth]{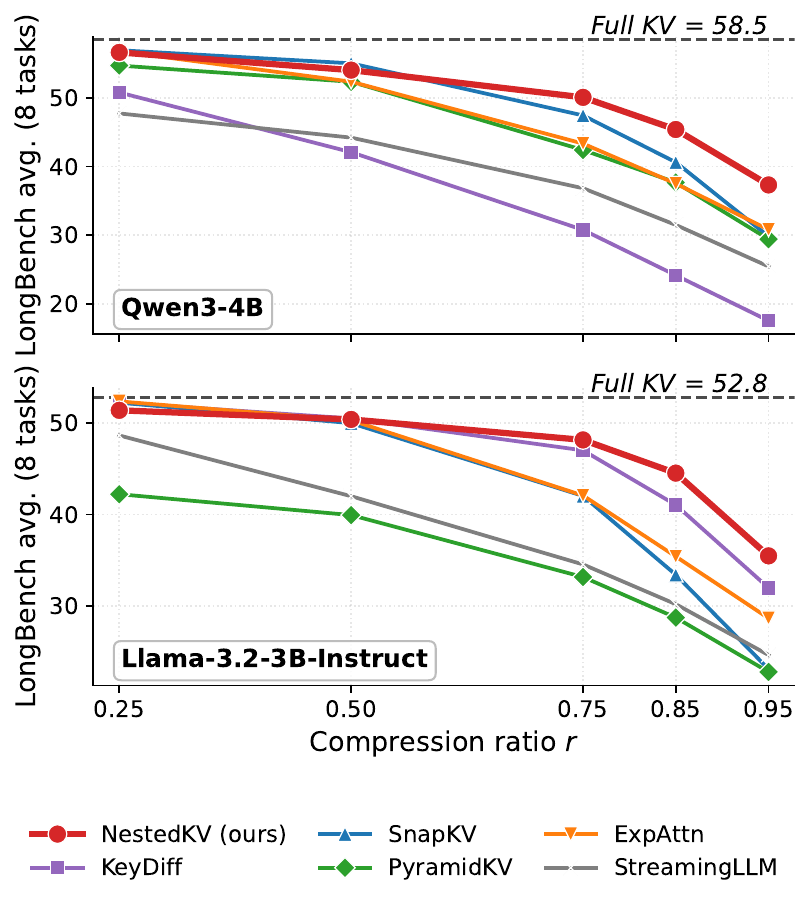}
  \caption{LongBench 8-task average vs.\ eviction ratio $r$ on Qwen3-4B (top) and Llama-3.2-3B-Instruct (bottom). Dashed lines mark each model's Full KV ceiling.}
  \label{fig:longbench-ratio}
\end{figure}

\subsection{Long-Context Tasks}

We organize the main results by benchmark. Each table or figure reports Full KV, multiple compression ratios, and multiple model families in a single view. NestedKV rows are shaded.

Table~\ref{tab:ruler-main} first reports RULER, a controlled synthetic benchmark that isolates retrieval and aggregation behaviour at 4k--32k contexts. Across Qwen3 and Llama-3.2-Instruct models, NestedKV is consistently strongest or near-strongest, and its advantage is clearest under aggressive compression. On Qwen3-4B, it improves over KeyDiff in all reported context--ratio cells; at $r=0.75$, the gains are +19.10 at 4k, +15.99 at 8k, and +17.69 at 16k. The 32k columns show the same trend at longer context: attention-based baselines can be competitive when the retained budget is still sufficient, but NestedKV avoids the sharp drops seen in single-signal methods as the budget tightens.

The Qwen3-32B results in Table~\ref{tab:qwen32-ruler} show the same pattern at a substantially larger scale. NestedKV is the highest-scoring compressed method in all six 4k/8k context--ratio configurations: for example, it reaches 95.10 versus 90.23 for KeyDiff at 4k/$r{=}0.25$, and 76.71 versus 71.17 at 8k/$r{=}0.75$.

\begin{table*}[!t]
\centering
\small
\setlength{\tabcolsep}{5pt}
\begin{tabular}{lrrrrrr}
\toprule
\multirow{2}{*}{Method} & \multicolumn{3}{c}{4k} & \multicolumn{3}{c}{8k} \\
\cmidrule(lr){2-4}\cmidrule(lr){5-7}
& $r=.25$ & $r=.50$ & $r=.75$ & $r=.25$ & $r=.50$ & $r=.75$ \\
\midrule
Full KV & & 95.73 & & & 94.90 & \\
PyramidKV & 67.56 & 34.88 & 36.64 & 66.68 & 36.82 & 28.41 \\
Expected Attention & 74.24 & 48.32 & 36.00 & 80.49 & 51.42 & 32.77 \\
SnapKV & 81.58 & 59.36 & 38.94 & 78.74 & 59.29 & 39.47 \\
StreamingLLM & 80.10 & 61.85 & 41.00 & 79.40 & 59.16 & 36.42 \\
KeyDiff & 90.23 & 79.67 & 69.38 & 90.25 & 78.85 & 71.17 \\
\rowcolor{nestedkvblue} NestedKV & \textbf{95.10} & \textbf{88.66} & \textbf{78.63} & \textbf{94.18} & \textbf{88.47} & \textbf{76.71} \\
\bottomrule
\end{tabular}
\caption{Qwen3-32B RULER average over 13 tasks. The shaded row is NestedKV; the best compressed-cache result in each context--ratio column is in bold.}
\label{tab:qwen32-ruler}
\end{table*}

Figure~\ref{fig:loogle-lines} evaluates real long-document QA on LooGLE. On long-dependency QA, NestedKV is best or within $0.6$ Rouge-L of the best method on three of four models, though PyramidKV is stronger on Qwen3-8B at $r\le 0.50$. On short-dependency QA, NestedKV leads on Qwen3-8B and Llama-3.2-3B-Instruct, while PyramidKV or ExpAttn can win in easier low-compression settings. The key trend is therefore not a uniform per-cell win, but stable performance across dependency regimes as compression increases.

Figure~\ref{fig:longbench-ratio} reports LongBench averages, where the main signal is the slope of degradation. At $r=0.25$ and $r=0.50$, the methods are close: on Qwen3-4B, SnapKV is marginally ahead of NestedKV and ExpAttn is competitive, while on Llama-3.2-3B-Instruct the strongest baseline differs by less than one point. The curves separate once the cache becomes scarce. On Qwen3-4B, NestedKV reaches 50.06 at $r=0.75$ versus 30.77 for KeyDiff, remains highest at $r=0.85$ (45.38 vs.\ 40.61 for SnapKV), and keeps 37.32 at $r=0.95$ while KeyDiff falls to 17.55. On Llama-3.2-3B-Instruct, NestedKV is similarly the strongest method from $r=0.75$ onward, reaching 48.14, 44.51, and 35.47 at $r\in\{0.75,0.85,0.95\}$. These results support the central claim: NestedKV is not primarily a low-compression peak-score method; its benefit is that multi-time-scale scoring degrades more gracefully when the retained cache must be small.

\subsection{Short-Context Preservation}
\label{sec:mmlu-pro}

\begin{figure}[t]
\centering
\includegraphics[width=\linewidth]{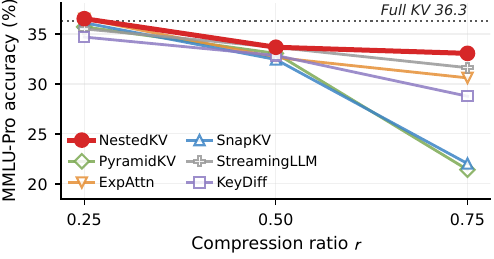}
\caption{MMLU-Pro accuracy on Qwen3-4B versus compression ratio $r$. The dotted line is the Full KV baseline.}
\label{fig:mmlu-pro}
\end{figure}

To check whether the long-context gains come at the cost of short-prompt capability, we evaluate MMLU-Pro \citep{wang2024mmlu} on Qwen3-4B in a $0$-shot setting. Figure~\ref{fig:mmlu-pro} reports accuracy versus the compression ratio $r$. NestedKV is the top or tied method at every ratio: essentially lossless at $r{=}0.25$ ($36.5$ vs Full KV $36.3$), and at $r{=}0.75$ it retains $33.1$ while SnapKV and PyramidKV collapse to $22.0$ and $21.4$. This indicates that continuum-memory scoring does not compromise short-prompt knowledge access.

\subsection{Ablation Study}
\label{sec:ablation}

We ablate NestedKV's two core components --- the continuum memory state and the adaptive head-wise allocation --- by removing them in isolation and jointly. We evaluate at Qwen3-4B RULER 4k under aggressive compression ($r=0.75$), where neither component is masked by ceiling effects.

Table~\ref{tab:ablation} reports the results. The two components contribute comparably: removing the adaptive head-wise budget drops the score by 8.41 points, while replacing the three-scale continuum score with a single-anchor key-distinctiveness score drops it by 7.99 points. Removing both jointly drops the score by 19.10 points, more than the sum of the individual deltas (16.40), because the two components are coupled by the discrete top-$k$ cache decision: continuum scoring decides \emph{which} tokens remain salient, while adaptive allocation decides \emph{where} the retained budget is spent across heads. Each component can partially compensate when the other remains, but removing both eliminates both compensation paths. We repeat the same four-variant ablation on LongBench and LooGLE in Appendix~\ref{app:ablation-cross} (Figure~\ref{fig:ablation-cross}); the same pattern holds but the continuum component becomes the dominant contributor on real-world long-document tasks. Table~\ref{tab:camera-ready-router} directly ablates surprise-guided routing, while Appendix~\ref{app:camera-ready} further isolates the stable, episodic, and current signals and compares three and four scales.

\begin{table}[t]
\centering
\small
\setlength{\tabcolsep}{4pt}
\begin{tabular}{lcc}
\toprule
Variant & RULER 4k, $r{=}0.75$ & $\Delta$ \\
\midrule
\rowcolor{nestedkvblue} NestedKV (full) & \textbf{79.32} & --- \\
w/o adaptive & 70.91 & $-8.41$ \\
w/o continuum & 71.33 & $-7.99$ \\
w/o adaptive \& continuum & 60.22 & $-19.10$ \\
\bottomrule
\end{tabular}
\caption{Component ablation of NestedKV on Qwen3-4B RULER 4k at $r=0.75$. ``w/o continuum'' replaces the three-scale continuum score with a single-anchor key-distinctiveness score; ``w/o adaptive'' replaces the head-adaptive budget with a uniform per-head budget. Both components contribute roughly equally; their combined removal exceeds the sum of the individual deltas.}
\label{tab:ablation}
\end{table}

\begin{table}[t]
\centering
\small
\setlength{\tabcolsep}{4pt}
\begin{tabular}{lc}
\toprule
Variant & RULER 4k, $r{=}0.75$ \\
\midrule
\rowcolor{nestedkvblue} NestedKV (Surprise Gate) & \textbf{79.32} \\
Only average & 76.25 \\
Only max & 75.54 \\
\bottomrule
\end{tabular}
\caption{Direct Surprise Gate ablation on Qwen3-4B RULER 4k at $r{=}0.75$. The shaded row is the full NestedKV routing rule; ``only average'' uses the blended score, and ``only max'' selects the highest per-scale score.}
\label{tab:camera-ready-router}
\end{table}

\begin{table}[t]
\centering
\small
\setlength{\tabcolsep}{4pt}
\begin{tabularx}{\columnwidth}{@{}l*{4}{>{\centering\arraybackslash}X}@{}}
\toprule
Method & KV & Prefill & Decode & Peak \\
\midrule
Full KV & 100\% & \textbf{6.28} & 38.90 & 21.48 \\
KeyDiff & 25\% & 6.34 & \textbf{31.46} & \textbf{18.11} \\
SnapKV  & 25\% & 6.50 & 31.49 & \textbf{18.11} \\
\rowcolor{nestedkvblue} NestedKV & 25\% & 6.36 & 31.47 & \textbf{18.11} \\
\bottomrule
\end{tabularx}
\caption{Efficiency on Qwen3-4B with a 32k-token context on a single NVIDIA L20 in bf16, $r{=}0.75$. KV denotes retained prompt-cache entries used during decoding; prefill is in seconds, decode in ms/token, and peak memory in GB.}
\label{tab:efficiency}
\end{table}

\subsection{Efficiency}

Table~\ref{tab:efficiency} reports prefill latency, per-token decode latency, and peak GPU memory of NestedKV against KeyDiff, SnapKV, and the Full KV baseline on Qwen3-4B with a 32k-token context. At the same $r$, all compressed methods use the same retained prompt-cache budget during decoding, which gives nearly identical decode latency and peak memory while remaining substantially below Full KV. The three-time-scale scoring incurs a one-time prefill overhead that is amortized over decoding, and stays within $0.5\%$ of the single-anchor KeyDiff prefill.

\section{Conclusion}

We presented NestedKV, a training-free KV cache compression method that treats eviction as bounded test-time memory maintenance. Instead of ranking cached tokens from a single signal, NestedKV uses three time-scale key statistics --- stable, episodic, and current --- as inner learners, each producing its own cosine anomaly ranking. A training-free outer learner combines these rankings on two axes: a head-adaptive softmax mix anchored by a fixed log-prior, and a per-token sigmoid gate over compression-induced surprise that routes between the blended view and the strongest individual memory. 

According to experiments, the results show that continuum-memory scoring is especially useful when single-anchor rules become brittle under stronger compression and longer contexts. The ablations further indicate that stable, episodic, and current memories contribute complementary signals, and that adaptive head-wise allocation amplifies these gains by matching memory budgets to head-specific residual structure. 


\section*{Limitations}

NestedKV assumes that redundancy with respect to stable, episodic, and current key statistics is a useful signal for cache eviction. This assumption is strongest for retrieval, question answering, and long-document understanding, where repeated local context often indicates information already represented by the continuum state. It can be weaker for code-completion style tasks: local repetition in code may be the very pattern the model must preserve for the next line. This suggests that future variants should adapt the continuum weights to input structure, for example, by reducing episodic or current redundancy penalties when the prompt exhibits code-like local regularity.

Our experiments are also limited to frozen open-weight models and training-free compression during prefill. NestedKV does not modify decoding kernels or learn task-specific parameters, which keeps the method simple but leaves open whether learned or query-aware variants could further improve difficult settings.

NestedKV is a prefill-only, query-agnostic compressor: it scores and prunes prompt KV entries once after prompt encoding, then performs standard autoregressive decoding. It does not recompute prompt-token scores, scale reliabilities, or routes as generation proceeds, and newly generated tokens are appended without compression. Decoding-time recompression is orthogonal to the present method and remains future work.

\section*{Acknowledgments}

We thank the reviewers for their valuable feedback and suggestions, which helped improve this work. This work was supported by the National Natural Science Foundation of China (Grant No.62506318); Guangdong Provincial Department of Education Project (Grant No.2024KQNCX028); Scientific Research Projects for the Higher-educational Institutions (Grant No.2024312096), Education Bureau of Guangzhou Municipality; Guangzhou-HKUST(GZ) Joint Funding Program (Grant No.2025A03J3957), Education Bureau of Guangzhou Municipality.

\bibliography{custom}

@article{liu2023scissorhands,
  title={Scissorhands: Exploiting the persistence of importance hypothesis for llm kv cache compression at test time},
  author={Liu, Zichang and Desai, Aditya and Liao, Fangshuo and Wang, Weitao and Xie, Victor and Xu, Zhaozhuo and Kyrillidis, Anastasios and Shrivastava, Anshumali},
  journal={Advances in Neural Information Processing Systems},
  volume={36},
  pages={52342--52364},
  year={2023}
}

@article{zhang2023h2o,
  title={H2o: Heavy-hitter oracle for efficient generative inference of large language models},
  author={Zhang, Zhenyu and Sheng, Ying and Zhou, Tianyi and Chen, Tianlong and Zheng, Lianmin and Cai, Ruisi and Song, Zhao and Tian, Yuandong and R{\'e}, Christopher and Barrett, Clark and others},
  journal={Advances in Neural Information Processing Systems},
  volume={36},
  pages={34661--34710},
  year={2023}
}

@inproceedings{xiao2024efficient,
  title={Efficient streaming language models with attention sinks},
  author={Xiao, Guangxuan and Tian, Yuandong and Chen, Beidi and Han, Song and Lewis, Mike},
  booktitle={International Conference on Learning Representations},
  volume={2024},
  pages={21875--21895},
  year={2024}
}

@article{li2024snapkv,
  title={Snapkv: Llm knows what you are looking for before generation},
  author={Li, Yuhong and Huang, Yingbing and Yang, Bowen and Venkitesh, Bharat and Locatelli, Acyr and Ye, Hanchen and Cai, Tianle and Lewis, Patrick and Chen, Deming},
  journal={Advances in Neural Information Processing Systems},
  volume={37},
  pages={22947--22970},
  year={2024}
}

@article{cai2024pyramidkv,
  title={Pyramidkv: Dynamic kv cache compression based on pyramidal information funneling},
  author={Cai, Zefan and Zhang, Yichi and Gao, Bofei and Liu, Yuliang and Li, Yucheng and Liu, Tianyu and Lu, Keming and Xiong, Wayne and Dong, Yue and Hu, Junjie and others},
  journal={arXiv preprint arXiv:2406.02069},
  year={2024}
}

@article{feng2026ada,
  title={Ada-kv: Optimizing kv cache eviction by adaptive budget allocation for efficient llm inference},
  author={Feng, Yuan and Lv, Junlin and Cao, Yukun and Xie, Xike and Zhou, S Kevin},
  journal={Advances in Neural Information Processing Systems},
  volume={38},
  pages={113152--113188},
  year={2026}
}

@article{park2026keydiff,
  title={Keydiff: Key similarity-based kv cache eviction for long-context llm inference in resource-constrained environments},
  author={Park, Junyoung and Jones, Dalton and Morse, Matthew and Goel, Raghavv and Lee, Mingu and Lott, Christopher},
  journal={Advances in Neural Information Processing Systems},
  volume={38},
  pages={5983--6019},
  year={2026}
}

@article{liu2025flowkv,
  title={FlowKV: Enhancing multi-turn conversational coherence in LLMs via isolated key-value cache management},
  author={Liu, Xiang and Chen, Hong and Hu, Xuming and Chu, Xiaowen},
  journal={arXiv preprint arXiv:2505.15347},
  year={2025}
}

@article{chen2026sonic,
  title={SONIC: Segmented Optimized Nexus for Information Compression in Key-Value Caching},
  author={Chen, Hong and Liu, Xiang and Wang, Bo and Fan, Yuxuan and Chu, Yuanlin and Li, Zongluo and Chu, Xiaowen and Hu, Xuming},
  journal={arXiv preprint arXiv:2601.21927},
  year={2026}
}

@misc{liu2026semanticintegritymattersbenchmarking,
      title={Semantic Integrity Matters: Benchmarking and Preserving High-Density Reasoning in KV Cache Compression}, 
      author={Xiang Liu and Zhenheng Tang and Hong Chen and Peijie Dong and Zeyu Li and Xiuze Zhou and Bo Li and Xuming Hu and Xiaowen Chu},
      year={2026},
      eprint={2502.01941},
      archivePrefix={arXiv},
      primaryClass={cs.CL},
      url={https://arxiv.org/abs/2502.01941}, 
}

@inproceedings{oren2024transformers,
  title={Transformers are multi-state rnns},
  author={Oren, Matanel and Hassid, Michael and Yarden, Nir and Adi, Yossi and Schwartz, Roy},
  booktitle={Proceedings of the 2024 Conference on Empirical Methods in Natural Language Processing},
  pages={18724--18741},
  year={2024}
}

@inproceedings{devoto2024simple,
  title={A simple and effective l\_2 norm-based strategy for kv cache compression},
  author={Devoto, Alessio and Zhao, Yu and Scardapane, Simone and Minervini, Pasquale},
  booktitle={Proceedings of the 2024 Conference on Empirical Methods in Natural Language Processing},
  pages={18476--18499},
  year={2024}
}

@article{devoto2025expected,
  title={Expected attention: Kv cache compression by estimating attention from future queries distribution},
  author={Devoto, Alessio and Jeblick, Maximilian and J{\'e}gou, Simon},
  journal={arXiv preprint arXiv:2510.00636},
  year={2025}
}

@article{behrouz2026nested,
  title={Nested learning: The illusion of deep learning architectures},
  author={Behrouz, Ali and Razaviyayn, Meisam and Zhong, Peilin and Mirrokni, Vahab},
  journal={Advances in Neural Information Processing Systems},
  volume={38},
  pages={46968--47002},
  year={2026}
}

@article{behrouz2026titans,
  title={Titans: Learning to memorize at test time},
  author={Behrouz, Ali and Zhong, Peilin and Mirrokni, Vahab},
  journal={Advances in Neural Information Processing Systems},
  volume={38},
  pages={113506--113543},
  year={2026}
}

@article{yang2025qwen3,
  title={Qwen3 technical report},
  author={Yang, An and Li, Anfeng and Yang, Baosong and Zhang, Beichen and Hui, Binyuan and Zheng, Bo and Yu, Bowen and Gao, Chang and Huang, Chengen and Lv, Chenxu and others},
  journal={arXiv preprint arXiv:2505.09388},
  year={2025}
}

@article{grattafiori2024llama,
  title={The llama 3 herd of models},
  author={Grattafiori, Aaron and Dubey, Abhimanyu and Jauhri, Abhinav and Pandey, Abhinav and Kadian, Abhishek and Al-Dahle, Ahmad and Letman, Aiesha and Mathur, Akhil and Schelten, Alan and Vaughan, Alex and others},
  journal={arXiv preprint arXiv:2407.21783},
  year={2024}
}

@article{hsieh2024ruler,
  title={RULER: What's the real context size of your long-context language models?},
  author={Hsieh, Cheng-Ping and Sun, Simeng and Kriman, Samuel and Acharya, Shantanu and Rekesh, Dima and Jia, Fei and Zhang, Yang and Ginsburg, Boris},
  journal={arXiv preprint arXiv:2404.06654},
  year={2024}
}

@inproceedings{bai2024longbench,
  title={Longbench: A bilingual, multitask benchmark for long context understanding},
  author={Bai, Yushi and Lv, Xin and Zhang, Jiajie and Lyu, Hongchang and Tang, Jiankai and Huang, Zhidian and Du, Zhengxiao and Liu, Xiao and Zeng, Aohan and Hou, Lei and others},
  booktitle={Proceedings of the 62nd annual meeting of the association for computational linguistics (volume 1: Long papers)},
  pages={3119--3137},
  year={2024}
}

@inproceedings{dasigi2021dataset,
  title={A dataset of information-seeking questions and answers anchored in research papers},
  author={Dasigi, Pradeep and Lo, Kyle and Beltagy, Iz and Cohan, Arman and Smith, Noah A and Gardner, Matt},
  booktitle={Proceedings of the 2021 Conference of the North American Chapter of the Association for Computational Linguistics: Human Language Technologies},
  pages={4599--4610},
  year={2021}
}

@inproceedings{li2024loogle,
  title={Loogle: Can long-context language models understand long contexts?},
  author={Li, Jiaqi and Wang, Mengmeng and Zheng, Zilong and Zhang, Muhan},
  booktitle={Proceedings of the 62nd Annual Meeting of the Association for Computational Linguistics (Volume 1: Long Papers)},
  pages={16304--16333},
  year={2024}
}

@article{zhang2024infty,
  title={Infty Bench: Extending Long Context Evaluation Beyond 100K Tokens},
  author={Zhang, Xinrong and Chen, Yingfa and Hu, Shengding and Xu, Zihang and Chen, Junhao and Hao, Moo Khai and Han, Xu and Thai, Zhen Leng and Wang, Shuo and Liu, Zhiyuan and others},
  journal={arXiv preprint arXiv:2402.13718},
  year={2024}
}

@article{wang2024mmlu,
  title={Mmlu-pro: A more robust and challenging multi-task language understanding benchmark},
  author={Wang, Yubo and Ma, Xueguang and Zhang, Ge and Ni, Yuansheng and Chandra, Abhranil and Guo, Shiguang and Ren, Weiming and Arulraj, Aaran and He, Xuan and Jiang, Ziyan and others},
  journal={Advances in Neural Information Processing Systems},
  volume={37},
  pages={95266--95290},
  year={2024}
}

\clearpage
\appendix

\section{Implementation Details}
\label{app:impl}

\paragraph{Scoring cost.}
NestedKV is training-free and parameter-free. The stable memory requires one mean over the key sequence; episodic memory is computed with block means; and current memory is computed with cumulative sums. The scoring cost is therefore $O(Nd)$ per layer and head. NestedKV only removes cached positions and leaves retained keys and values unchanged, so the compressed memory remains compatible with standard attention implementations.

\paragraph{Inference pipeline.}
All methods are evaluated using the same \texttt{kvpress}-based runner and the same HuggingFace model checkpoints. NestedKV is implemented as a training-free prefill compressor: after the prompt is encoded, each layer computes per-scale anomalies from cached keys, combines them with the head-adaptive blend and surprise-gated route described in Section~\ref{sec:surprise}, allocates head-wise memory budgets, and removes low-scoring entries before decoding. The head-adaptive blend weights and surprise gates are computed once during this prefill-time compression step and then kept fixed throughout autoregressive decoding. NestedKV does not recompute scores, scale reliabilities, or routes for retained prompt tokens as new tokens are generated; newly decoded tokens are appended normally and remain available to subsequent decoding steps. Adapting the continuum-memory weights during very long-form generation is left to future work.

\paragraph{Hyperparameters.}
Unless otherwise stated, NestedKV uses log-prior $(w_s^0,w_e^0,w_c^0)=(0.4,0.4,0.2)$, block size $\operatorname{clip}(\lfloor N/32\rfloor,128,256)$, current-window size $64$, and $4$ pinned sink tokens. The head-adaptive blend uses $\beta=3.0$; the surprise gate uses threshold $\tau=0.60$ and sharpness $\kappa=10.0$. Before applying the gate, surprise scores are min--max normalized within each head and mean-centered with a rectifier. For head-wise budgeting, we use a per-head safeguard $\alpha_s=0.20$: for a sequence of length $N$ and eviction ratio $r$, each head first preserves $\lfloor 0.20\lfloor(1-r)N\rfloor\rfloor$ of its own highest-scoring tokens before the remaining budget is allocated by cross-head competition. These constants are fixed across all models, benchmarks, and compression ratios.

\begin{table}[t]
\centering
\small
\setlength{\tabcolsep}{3pt}
\begin{tabularx}{\columnwidth}{@{}lXrr@{}}
\toprule
Benchmark & Scope & Examples & Avg. toks. \\
\midrule
RULER & 13 tasks, 4k--32k & 6,500 & 19.1k \\
LongBench & 8 tasks & 1,550 & 8.6k \\
LongBench-E & 13 tasks & 3,668 & 5.8k \\
LooGLE & long/short QA & 3,052 & 25.6k \\
InfiniteBench & longbook+code & 745 & 40.0k \\
MMLU-Pro & 10-choice & 12,032 & 232 \\
\bottomrule
\end{tabularx}
\caption{Evaluation data scale. ``Examples'' counts the fixed evaluation set per model. Average input length is measured in tokens under the evaluation prompt construction; RULER evaluates 6,500 samples under per context length setting, and InfiniteBench reflects the $40$k input cap.}
\label{tab:dataset-statistics}
\end{table}

\section{Additional Analyses}
\label{app:camera-ready}

\subsection{\texorpdfstring{Answer-Span Retention}{Answer-Span Retention}}

\begin{table}[t]
\centering
\small
\setlength{\tabcolsep}{4pt}
\begin{tabular}{lrrrr}
\toprule
Method & All & Needle/retrieval & QA & \texttt{fwe}/\texttt{vt} \\
\midrule
Attention-only & 68.97 & 48.23 & \textbf{74.60} & \textbf{100.00} \\
\rowcolor{nestedkvblue} NestedKV & \textbf{92.29} & \textbf{97.35} & 47.62 & \textbf{100.00} \\
\bottomrule
\end{tabular}
\caption{Answer-span retention (\%) at the same top-$k$ budget on Qwen3-4B RULER 4k at $r{=}0.75$. The shaded row is NestedKV; per-subset maxima are in bold.}
\label{tab:answer-retention}
\end{table}

\subsection{\texorpdfstring{Component and Scale Ablations}{Component and Scale Ablations}}

The following ablation separates the contribution of the multi-scale memory score from head-wise memory allocation. A single stable-memory score, removing current memory, and removing head-wise allocation all reduce the RULER score relative to the full method.

\begin{table}[t]
\centering
\small
\setlength{\tabcolsep}{3pt}
\begin{tabular}{lr}
\toprule
Variant & Average \\
\midrule
Single-scale stable score & 71.58 \\
Stable + episodic (no current) & 74.81 \\
Multi-scale score only (no allocation) & 71.48 \\
NestedKV & 79.32 \\
\bottomrule
\end{tabular}
\caption{Component ablation on Qwen3-4B RULER 4k at $r{=}0.75$. The single-scale variant uses the same allocation mechanism as NestedKV.}
\label{tab:camera-ready-components}
\end{table}

Current memory is therefore a complementary local signal rather than a component expected to improve every task monotonically. In a direct scale-count check, adding a fourth 256-token local window does not improve the three-scale adaptive blend (77.94 versus 78.09), supporting three scales as an effectiveness--complexity trade-off rather than a theoretically unique choice.

\subsection{\texorpdfstring{Attention-History Feasibility Study}{Attention-History Feasibility Study}}

We separately evaluate ObservedAttention as an H$_2$O-like attention-history baseline. Rather than assigning a score of zero to failed runs, we report the observed OOM frequency. The method is weak on the runnable RULER setting and becomes impractical on the longer-context benchmarks in our eager-attention setup.

\begin{table}[t]
\centering
\scriptsize
\setlength{\tabcolsep}{3pt}
\begin{tabular}{lrr}
\toprule
Benchmark & Result & OOM frequency \\
\midrule
RULER 4k & 26.07 & 0/200 \\
LongBench & 19.15 & 111/200 \\
LongBench-E & 23.51 & 96/200 \\
LooGLE & N/A & 200/200 \\
MMLU-Pro & 37.24\% & 0/196 \\
InfiniteBench longbook & N/A & 100/100 \\
InfiniteBench code\_debug & N/A & 100/100 \\
\bottomrule
\end{tabular}
\caption{ObservedAttention feasibility study on Qwen3-4B at $r{=}0.75$.}
\label{tab:observed-attention}
\end{table}

\subsection{\texorpdfstring{Design-Choice Sensitivity}{Design-Choice Sensitivity}}

The default episodic schedule is $N/32$ clipped to $[128,256]$, the current-memory window is 64 tokens, and four sink tokens are pinned. The existing schedule and router sensitivity analyses above are complemented by the sink-token sweep below: four sink tokens capture nearly all of the observed gain, and larger sink budgets provide no material improvement.

\begin{table}[t]
\centering
\small
\setlength{\tabcolsep}{4pt}
\begin{tabular}{lr}
\toprule
Pinned sink tokens & Average \\
\midrule
0 & 74.59 \\
4 (default) & 79.32 \\
8 & 75.87 \\
16 & 75.87 \\
\bottomrule
\end{tabular}
\caption{Sink-token sensitivity on Qwen3-4B RULER 4k at $r{=}0.75$.}
\label{tab:sink-sensitivity}
\end{table}

\begin{table*}[t]
\centering
\small
\setlength{\tabcolsep}{3pt}
\renewcommand{\arraystretch}{0.88}
\resizebox{0.8\textwidth}{!}{%
\begin{tabular}{@{}clccccccccc@{}}
\toprule
Ratio & Method & NarrQA & Qasper & MF-QA & HotpotQA & 2Wiki & GovRpt & TrivQA & PsgRet & Avg \\
\midrule
-- & Full KV 
& 31.45 & 42.93 & 54.82 & 68.05 
& 55.05 & 30.00 & 85.89 & 100.00 & 58.52 \\
\midrule

\multirow{6}{*}{$r{=}0.25$}
& PyramidKV
& 26.09 & 40.05 & 49.99 & 63.15
& 55.65 & 29.16 & 80.11 & 93.33 & 54.69 \\
& ExpAttn
& 27.87 & 40.10 & 51.89 & 67.42
& 53.83 & 30.51 & 85.89 & 96.67 & 56.77 \\
& SnapKV
& 27.02 & 37.47 & 54.18 & 67.86
& 53.52 & 30.57 & 86.44 & 98.33 & 56.92 \\
& StreamingLLM
& 24.28 & 36.66 & 33.60 & 55.26
& 42.99 & 27.81 & 79.52 & 81.67 & 47.72 \\
& KeyDiff
& 25.60 & 34.06 & 45.90 & 48.71
& 45.10 & 29.81 & 83.68 & 93.33 & 50.77 \\
\rowcolor{nestedkvblue}
& NestedKV
& 30.07 & 40.58 & 50.79 & 64.19
& 51.85 & 29.47 & 87.56 & 98.33 & 56.60 \\
\midrule

\multirow{6}{*}{$r{=}0.50$}
& PyramidKV
& 24.21 & 33.50 & 31.82 & 56.76
& 46.07 & 26.38 & 80.11 & 91.67 & 48.82 \\
& ExpAttn
& 28.17 & 38.55 & 46.82 & 63.31
& 48.89 & 30.45 & 87.56 & 75.00 & 52.34 \\
& SnapKV
& 29.45 & 36.32 & 41.48 & 65.41
& 53.29 & 29.38 & 86.44 & 98.33 & 55.01 \\
& StreamingLLM
& 23.80 & 30.25 & 33.56 & 53.98
& 47.00 & 28.39 & 76.63 & 60.00 & 44.20 \\
& KeyDiff
& 26.56 & 25.85 & 39.69 & 40.04
& 32.19 & 25.00 & 77.33 & 70.00 & 42.08 \\
\rowcolor{nestedkvblue}
& NestedKV
& 28.40 & 38.77 & 42.77 & 63.67
& 45.34 & 27.37 & 87.56 & 98.33 & 54.03 \\
\midrule

\multirow{6}{*}{$r{=}0.75$}
& PyramidKV
& 21.31 & 29.52 & 25.26 & 42.51
& 38.31 & 24.17 & 81.44 & 76.67 & 42.40 \\
& ExpAttn
& 22.28 & 37.08 & 41.86 & 59.25
& 48.13 & 28.70 & 85.18 & 24.17 & 43.33 \\
& SnapKV
& 21.59 & 29.25 & 28.87 & 60.35
& 39.55 & 26.72 & 86.44 & 86.67 & 47.43 \\
& StreamingLLM
& 21.22 & 24.93 & 29.55 & 45.43
& 40.72 & 26.31 & 74.85 & 31.67 & 36.84 \\
& KeyDiff
& 19.51 & 12.18 & 28.07 & 25.86
& 26.16 & 17.34 & 70.39 & 46.67 & 30.77 \\
\rowcolor{nestedkvblue}
& NestedKV
& 27.21 & 30.40 & 39.52 & 54.04
& 39.87 & 25.23 & 85.89 & 98.33 & 50.06 \\
\midrule

\multirow{6}{*}{$r{=}0.85$}
& PyramidKV
& 20.32 & 20.28 & 29.40 & 36.84
& 31.26 & 23.34 & 85.45 & 54.50 & 37.67 \\
& ExpAttn
& 23.14 & 30.51 & 34.98 & 49.19
& 33.54 & 28.30 & 86.26 & 14.17 & 37.51 \\
& SnapKV
& 22.82 & 20.79 & 31.28 & 43.97
& 30.33 & 24.87 & 86.85 & 64.00 & 40.61 \\
& StreamingLLM
& 18.86 & 19.72 & 27.10 & 37.11
& 30.35 & 24.48 & 72.84 & 21.50 & 31.50 \\
& KeyDiff
& 13.92 & 11.51 & 22.11 & 19.52
& 24.04 & 13.84 & 69.89 & 18.50 & 24.17 \\
\rowcolor{nestedkvblue}
& NestedKV
& 23.99 & 23.42 & 33.67 & 46.85
& 27.54 & 23.78 & 86.82 & 97.00 & 45.38 \\
\midrule

\multirow{6}{*}{$r{=}0.95$}
& PyramidKV
& 15.04 & 13.06 & 24.01 & 34.01
& 24.83 & 19.48 & 86.40 & 18.58 & 29.43 \\
& ExpAttn
& 18.60 & 22.29 & 26.22 & 37.42
& 22.68 & 25.82 & 83.44 & 10.00 & 30.81 \\
& SnapKV
& 18.00 & 13.18 & 23.78 & 33.80
& 24.78 & 19.58 & 86.73 & 18.75 & 29.82 \\
& StreamingLLM
& 17.36 & 14.17 & 22.90 & 31.09
& 23.83 & 19.41 & 66.37 & 8.50 & 25.45 \\
& KeyDiff
& 9.43 & 3.97 & 17.79 & 9.11
& 22.92 & 9.17 & 64.01 & 4.00 & 17.55 \\
\rowcolor{nestedkvblue}
& NestedKV
& 18.83 & 16.02 & 25.04 & 32.97
& 25.95 & 19.40 & 85.32 & 75.00 & 37.32 \\
\bottomrule
\end{tabular}}
\caption{LongBench per-task scores on Qwen3-4B across different KV eviction ratios. NestedKV rows are shaded.}
\label{tab:longbench-detail-qwen34b}
\end{table*}

\section{Related Work}

\paragraph{Training-free KV cache eviction and budgeting.}
KV cache compression reduces inference memory by retaining only a subset of past keys and values. A central line of work ranks cached tokens by attention-derived importance. Scissorhands proposes the persistence-of-importance hypothesis, observing that tokens important in previous attention tend to remain important later \citep{liu2023scissorhands}. H$_2$O similarly keeps heavy hitters in attention scores while preserving recent tokens \citep{zhang2023h2o}. StreamingLLM identifies the importance of initial attention sinks and combines them with a recent window to support streaming generation \citep{xiao2024efficient}. SnapKV uses an observation window at the end of the prompt to estimate which prefix tokens are likely to matter during generation \citep{li2024snapkv}. Expected Attention estimates future attention scores from the distribution of future queries rather than directly observing them \citep{devoto2025expected}. Ada-KV addresses a different but related issue: after tokens are scored, the available cache budget should be allocated adaptively across heads rather than uniformly \citep{feng2026ada}. NestedKV follows this line of training-free compression, but defines token importance as a multi-time-scale anomaly of the stored key stream rather than as a single attention, recency, or head-allocation statistic.

\paragraph{Geometric, structured, and conversational cache compression.}
Other methods avoid direct attention scoring and compress the cache using structure in key/value representations or dialogue history. KeyDiff ranks tokens by their cosine difference from the mean key direction, showing that geometric distinctiveness is a strong training-free signal for long-context inference \citep{park2026keydiff}. L$_2$-norm-based compression uses the magnitude of key embeddings as a simple importance proxy \citep{devoto2024simple}. PyramidKV studies layer-wise differences in cache importance and allocates memory according to a pyramidal information-funneling pattern \citep{cai2024pyramidkv}. FlowKV studies multi-turn conversations and isolates the KV cache of completed turns to avoid repeatedly recompressing older conversational context \citep{liu2025flowkv}. SONIC further emphasizes segment-level structure by compressing historical dialogue segments into compact Nexus tokens \citep{chen2026sonic}. Beyond efficiency, KVFundaBench shows that compression can affect fundamental abilities such as world knowledge, reasoning, code generation, safety, and long-context generation in task-dependent ways \citep{liu2026semanticintegritymattersbenchmarking}. NestedKV is closest in spirit to key-space geometric methods, but replaces the single global anchor with a continuum anchor spanning global, block-local, and recent statistics.

\paragraph{Nested learning and test-time memory.}
Several recent works reinterpret Transformer inference as maintaining a bounded recurrent or memory state rather than as unbounded full-context attention. StreamingLLM emphasizes attention sinks as persistent state variables for streaming generation \citep{xiao2024efficient}, while TOVA explicitly connects KV cache compression to bounded multi-state RNNs \citep{oren2024transformers}. Titans introduces a neural long-term memory module that learns to memorize at test time, contrasting attention as short-term memory with a persistent learned memory state \citep{behrouz2026titans}. Nested Learning further frames models as nested optimization problems over context flows and presents HOPE with a Continuum Memory System spanning multiple memory time scales and a self-modifying update rule that adapts to compression-induced surprise \citep{behrouz2026nested}. NestedKV instantiates both ideas for frozen LLM KV compression. The continuum-memory view drives the inner learners: each cached key is read against stable, episodic, and current memory anchors, producing three anomaly rankings rather than one. The self-modifying view drives the outer learner: cross-head reliability gaps and per-token cross-scale disagreement determine, respectively, how the three rankings are mixed on each head and whether the score falls back to the strongest individual memory on each token. Neither axis introduces trainable parameters or modifies the underlying LLM.

\section{Additional Results}
\label{app:additional}

\subsection{LongBench}
\label{app:longbench-detail}

Tables~\ref{tab:longbench-detail-qwen34b} and Tables~\ref{tab:longbench-detail-llama323b} report per-task scores for the eight LongBench tasks used in Figure~\ref{fig:longbench-ratio}, covering both models across all evaluated eviction ratios. Task abbreviations: NarrQA = NarrativeQA, MF-QA = MultiFieldQA-en, 2Wiki = 2WikiMQA, GovRpt = GovReport, TrivQA = TriviaQA, PsgRet = Passage-Retrieval-en.

\begin{table*}[ht]
\centering
\small
\setlength{\tabcolsep}{3pt}
\renewcommand{\arraystretch}{0.88}
\resizebox{0.8\textwidth}{!}{%
\begin{tabular}{@{}clccccccccc@{}}
\toprule
Ratio & Method & NarrQA & Qasper & MF-QA & HotpotQA & 2Wiki & GovRpt & TrivQA & PsgRet & Avg \\
\midrule
-- & Full KV
& 29.06 & 39.04 & 52.91 & 52.26
& 50.57 & 32.79 & 69.11 & 96.67 & 52.80 \\
\midrule

\multirow{6}{*}{$r{=}0.25$}
& PyramidKV
& 25.65 & 34.98 & 45.79 & 46.62
& 47.27 & 30.79 & 73.23 & 33.33 & 42.21 \\
& ExpAttn
& 29.75 & 36.63 & 53.76 & 50.75
& 50.84 & 31.01 & 69.39 & 96.67 & 52.35 \\
& SnapKV
& 28.15 & 38.35 & 53.75 & 53.37
& 49.86 & 31.81 & 67.44 & 95.00 & 52.22 \\
& StreamingLLM
& 24.29 & 36.84 & 35.34 & 53.16
& 43.39 & 30.19 & 87.48 & 78.33 & 48.63 \\
& KeyDiff
& 29.36 & 38.79 & 55.26 & 52.08
& 51.12 & 31.69 & 64.44 & 95.00 & 52.22 \\
\rowcolor{nestedkvblue}
& NestedKV
& 28.39 & 38.13 & 54.36 & 53.11
& 50.21 & 31.80 & 58.33 & 96.67 & 51.38 \\
\midrule

\multirow{6}{*}{$r{=}0.50$}
& PyramidKV
& 25.28 & 30.92 & 40.70 & 42.39
& 48.49 & 26.67 & 78.28 & 26.67 & 39.93 \\
& ExpAttn
& 30.51 & 38.18 & 47.84 & 52.00
& 51.34 & 30.50 & 63.56 & 88.33 & 50.28 \\
& SnapKV
& 25.67 & 37.80 & 45.70 & 54.40
& 42.29 & 29.80 & 69.11 & 95.00 & 49.97 \\
& StreamingLLM
& 20.67 & 30.69 & 25.74 & 50.41
& 40.28 & 28.97 & 84.15 & 55.00 & 41.99 \\
& KeyDiff
& 31.85 & 37.44 & 52.10 & 51.39
& 41.03 & 30.65 & 64.44 & 95.00 & 50.49 \\
\rowcolor{nestedkvblue}
& NestedKV
& 27.32 & 38.67 & 50.50 & 54.95
& 43.82 & 31.66 & 59.44 & 96.67 & 50.38 \\
\midrule

\multirow{6}{*}{$r{=}0.75$}
& PyramidKV
& 22.82 & 21.94 & 21.88 & 38.31
& 43.45 & 25.24 & 74.94 & 16.67 & 33.16 \\
& ExpAttn
& 28.29 & 30.60 & 37.42 & 57.56
& 47.02 & 29.11 & 68.19 & 38.33 & 42.06 \\
& SnapKV
& 24.10 & 26.97 & 26.91 & 47.52
& 37.07 & 26.12 & 65.78 & 81.67 & 42.02 \\
& StreamingLLM
& 15.57 & 24.02 & 22.30 & 40.58
& 36.20 & 25.97 & 84.95 & 26.67 & 34.53 \\
& KeyDiff
& 30.78 & 29.80 & 42.01 & 45.48
& 40.21 & 27.71 & 66.67 & 93.33 & 47.00 \\
\rowcolor{nestedkvblue}
& NestedKV
& 29.03 & 37.79 & 46.88 & 48.82
& 38.11 & 30.04 & 57.78 & 96.67 & 48.14 \\
\midrule

\multirow{6}{*}{$r{=}0.85$}
& PyramidKV
& 19.64 & 20.29 & 23.15 & 31.41
& 25.16 & 23.87 & 69.77 & 16.50 & 28.72 \\
& ExpAttn
& 24.21 & 28.17 & 32.14 & 46.42
& 29.58 & 29.02 & 69.18 & 24.50 & 35.40 \\
& SnapKV
& 20.90 & 20.56 & 24.68 & 38.50
& 25.68 & 25.00 & 44.93 & 67.00 & 33.41 \\
& StreamingLLM
& 18.28 & 19.02 & 20.13 & 30.85
& 23.70 & 25.29 & 84.74 & 19.50 & 30.19 \\
& KeyDiff
& 23.33 & 22.18 & 33.65 & 44.06
& 27.38 & 27.52 & 65.67 & 84.50 & 41.04 \\
\rowcolor{nestedkvblue}
& NestedKV
& 23.31 & 31.21 & 42.26 & 47.29
& 30.81 & 29.57 & 57.64 & 94.00 & 44.51 \\
\midrule

\multirow{6}{*}{$r{=}0.95$}
& PyramidKV
& 13.00 & 11.75 & 17.47 & 30.35
& 16.69 & 21.11 & 55.42 & 16.50 & 22.79 \\
& ExpAttn
& 19.09 & 21.08 & 21.13 & 36.90
& 19.50 & 26.32 & 79.22 & 6.00 & 28.66 \\
& SnapKV
& 16.94 & 11.15 & 17.37 & 30.34
& 16.35 & 21.18 & 54.92 & 16.50 & 23.09 \\
& StreamingLLM
& 13.82 & 14.66 & 15.02 & 23.67
& 21.18 & 21.56 & 80.16 & 7.00 & 24.63 \\
& KeyDiff
& 20.69 & 13.68 & 24.44 & 29.34
& 19.92 & 23.96 & 69.71 & 54.00 & 31.97 \\
\rowcolor{nestedkvblue}
& NestedKV
& 20.91 & 18.73 & 24.63 & 32.89
& 24.99 & 26.27 & 59.33 & 76.00 & 35.47 \\
\bottomrule
\end{tabular}}
\caption{LongBench per-task scores on Llama-3.2-3B-Instruct across different KV eviction ratios. NestedKV rows are shaded.}
\label{tab:longbench-detail-llama323b}
\end{table*}

\subsection{LongBench-E}
\label{app:longbench-e}

Table~\ref{tab:longbench-e-appendix} reports LongBench-E results, the length-balanced variant of LongBench with 0--4k, 4--8k, and 8k+ input buckets, averaged over its 13 tasks. On the smaller Llama-3.2-3B-Instruct model, NestedKV is within a fraction of a point of the best baseline at $r\le 0.50$, trails KeyDiff by 2.2 points at $r=0.75$, and is the top method at $r\in\{0.85, 0.95\}$. On Qwen3-4B, NestedKV stays within roughly four points of the best attention-based baseline at $r\le 0.85$ and becomes the top method at $r=0.95$. The pattern is consistent with the main text: continuum-memory scoring is most useful under aggressive compression, while at moderate compression all attention- and key-based methods cluster tightly.

\begin{table}[ht]
\centering
\small
\setlength{\tabcolsep}{4pt}
\begin{tabular}{lccccc}
\toprule
Method & $r{=}.25$ & $r{=}.50$ & $r{=}.75$ & $r{=}.85$ & $r{=}.95$ \\
\midrule
\multicolumn{6}{l}{\textit{Qwen3-4B} (Full KV $=48.77$)} \\
PyramidKV    & 47.50 & 43.07 & 41.20 & 37.76 & 31.39 \\
ExpAttn      & 48.92 & 47.53 & 43.58 & 45.11 & 31.95 \\
SnapKV       & 48.45 & 47.35 & 44.26 & 40.02 & 31.47 \\
StreamingLLM & 44.07 & 40.01 & 35.32 & 33.57 & 28.33 \\
KeyDiff      & 44.97 & 40.38 & 29.73 & 23.47 & 17.52 \\
\rowcolor{nestedkvblue} NestedKV & 48.69 & 46.75 & 43.68 & 41.53 & 34.89 \\
\midrule
\multicolumn{6}{l}{\textit{Llama-3.2-3B-Instruct} (Full KV $=39.21$)} \\
PyramidKV    & 41.10 & 37.21 & 33.20 & 30.76 & 24.15 \\
ExpAttn      & 40.11 & 41.15 & 37.20 & 32.81 & 24.75 \\
SnapKV       & 38.80 & 37.01 & 33.95 & 30.24 & 24.14 \\
StreamingLLM & 37.95 & 33.56 & 29.02 & 27.14 & 24.42 \\
KeyDiff      & 42.50 & 41.53 & 40.78 & 34.42 & 27.17 \\
\rowcolor{nestedkvblue} NestedKV & 42.28 & 41.15 & 38.57 & 36.10 & 28.99 \\
\bottomrule
\end{tabular}
\caption{LongBench-E 13-task average across compression ratios. Shaded rows are NestedKV.}
\label{tab:longbench-e-appendix}
\end{table}

\subsection{InfiniteBench longbook\_qa\_eng}
\label{app:infinitebench}

\begin{table}[ht]
\centering
\small
\setlength{\tabcolsep}{6pt}
\begin{tabular}{lccc}
\toprule
Method & $r{=}.25$ & $r{=}.50$ & $r{=}.75$ \\
\midrule
\multicolumn{4}{l}{\textit{Qwen3-4B} (Full KV $=12.09$)} \\
KeyDiff      & 10.98 & 9.63 & 7.30 \\
SnapKV       & 10.13 & 9.47 & 7.77 \\
PyramidKV    & 10.76 & 10.40 & 10.19 \\
ExpAttn      & 10.00 & 9.06 & 6.79 \\
StreamingLLM & 5.40 & 5.11 & 4.81 \\
\rowcolor{nestedkvblue} NestedKV & 11.20 & 10.58 & 11.20 \\
\bottomrule
\end{tabular}
\caption{InfiniteBench longbook\_qa\_eng F1 score (\%) on Qwen3-4B over all $n{=}351$ examples with $40$k input cap. Shaded row is NestedKV; per-column best in bold.}
\label{tab:infinitebench-appendix}
\end{table}

Table~\ref{tab:infinitebench-appendix} reports the long-document QA setting of InfiniteBench~\citep{zhang2024infty}, the longbook\_qa\_eng task, in which the model must answer free-form questions over book-length English contexts truncated to $40$k tokens. Absolute F1 is low for all methods ($\sim$10\%) because Qwen3-4B is itself a weak book-length QA model under token-level F1 with $1$--$3$ word gold answers. The informative signal is robustness as the retained cache shrinks. At light and moderate compression ($r{=}.25,.50$), all methods except StreamingLLM remain within roughly three points of the Full KV baseline ($12.09$), so no method is meaningfully separated. The separation appears only at aggressive compression: at $r{=}.75$ NestedKV ($11.20$) is the best method and is the only one that does not degrade --- its score is unchanged from $r{=}.25$ --- while KeyDiff, SnapKV, and ExpAttn fall to the $6.8$--$7.8$ range. PyramidKV is the only baseline that remains comparably robust ($10.19$). The combination of stable, episodic, and current anchors keeps a more diverse set of supporting tokens than a single-anchor scorer, which is what preserves usable accuracy once the budget becomes tight.

\subsection{InfiniteBench code\_debug}
\label{app:infinitebench-code-debug}

\begin{table}[ht]
\centering
\small
\setlength{\tabcolsep}{6pt}
\begin{tabular}{lccc}
\toprule
Method & $r{=}.25$ & $r{=}.50$ & $r{=}.75$ \\
\midrule
\multicolumn{4}{l}{\textit{Qwen3-4B} (Full KV $=34.26$)} \\
KeyDiff      & 31.47 & 32.23 & 30.96 \\
SnapKV       & 32.49 & 30.20 & 28.17 \\
PyramidKV    & 26.40 & 23.86 & 26.65 \\
ExpAttn      & 31.98 & 30.46 & 28.17 \\
StreamingLLM & 32.49 & 31.98 & 29.44 \\
\rowcolor{nestedkvblue} NestedKV & 33.76 & 34.01 & 31.22 \\
\bottomrule
\end{tabular}
\caption{InfiniteBench code\_debug accuracy (\%) on Qwen3-4B over all $n{=}394$ examples with $40$k input cap. Shaded row is NestedKV; per-column best in bold.}
\label{tab:infinitebench-code-debug-appendix}
\end{table}

Table~\ref{tab:infinitebench-code-debug-appendix} complements the MMLU-Pro analysis in Section~\ref{sec:mmlu-pro} with a code-intensive multiple-choice task from InfiniteBench~\citep{zhang2024infty}. The model must identify which function in a repository-scale Python context contains a deliberate bug. Because the answer is a single multiple-choice label rather than a free-form span, all methods are far more robust to compression here than on longbook\_qa\_eng: every method stays within a few points of the Full KV baseline ($34.26$) across all three ratios. NestedKV is nonetheless the strongest method at every ratio ($33.76$, $34.01$, $31.22$), slightly exceeding even Full KV, while PyramidKV is the weakest ($23.9$--$26.7$). This indicates that the continuum-memory score preserves compact code-level answer selection at least as well as any single-signal baseline, and does so consistently as the budget tightens.

\subsection{Cross-Benchmark Ablation}
\label{app:ablation-cross}

\begin{figure}[ht]
  \centering
  \includegraphics[width=\columnwidth]{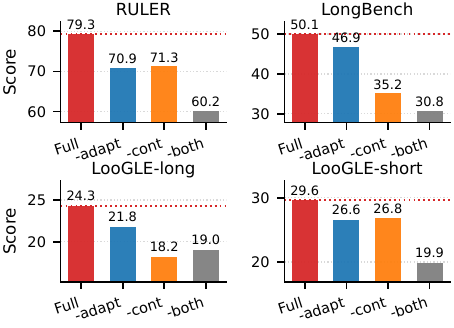}
  \caption{Cross-benchmark ablation on Qwen3-4B at $r=0.75$. Red bars show full NestedKV; other bars remove adaptive budgeting, continuum scoring, or both.}
  \label{fig:ablation-cross}
\end{figure}


Figure~\ref{fig:ablation-cross} extends the RULER 4k ablation of Section~\ref{sec:ablation} to LongBench and LooGLE. The qualitative picture is consistent across the four benchmarks: NestedKV (full) is the top configuration on every column, and removing both components is the worst on three of the four. However, the relative contribution of the two components depends strongly on the task. On RULER 4k the two components contribute almost equally ($\!-8.41$ vs $\!-7.99$), but on LongBench and LooGLE-long the continuum score is the dominant factor: $-14.89$ vs $-3.18$ on LongBench and $-6.04$ vs $-2.47$ on LooGLE-long, i.e.\ continuum scoring carries between $2.4\times$ and $4.7\times$ the per-task value of head-adaptive allocation. On LooGLE-short, the two components are again comparable and small individually, but jointly account for $-9.67$ points -- the strongest super-additivity among the ablations. This pattern suggests that the controlled synthetic structure of RULER masks how much of NestedKV's gain on real-world long-document tasks is driven by the continuum-memory score, rather than by adaptive head-wise allocation alone.

\subsection{Hyperparameter Sensitivity}
\label{app:sensitivity}

\begin{figure}[ht]
  \centering
  \includegraphics[width=\columnwidth]{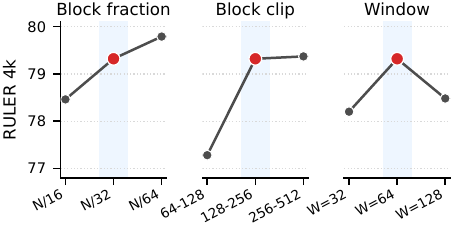}
  \caption{Hyperparameter sensitivity on Qwen3-4B RULER 4k at $r=0.75$. Red markers indicate the default schedule used in the main results.}
  \label{fig:sensitivity}
\end{figure}


Figure~\ref{fig:sensitivity} reports the RULER 4k score at $r=0.75$ as each block/window schedule knob is moved one step away from the default. Across the three axes the scores stay within $2.04$ points of the default; the smallest block clip ($[64,128]$) is the only configuration losing more than $1.2$ points, and the largest block fraction ($N/64$) is marginally better than the default ($+0.46$, within run-to-run noise). NestedKV is therefore robust to the precise block and window schedule in the explored neighbourhood, and the default schedule sits within $0.47$ points of the best observed configuration.

\begin{figure}[ht]
  \centering
  \includegraphics[width=\columnwidth]{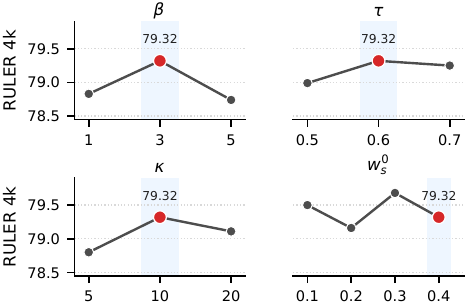}
  \caption{Router/prior hyperparameter sensitivity on Qwen3-4B RULER 4k at $r=0.75$. Red markers indicate the default configuration used in the main results.}
  \label{fig:router-sensitivity}
\end{figure}


Figure~\ref{fig:router-sensitivity} reports the same robustness probe for the four router and prior knobs introduced in Section~\ref{sec:surprise}: the mixing temperature $\beta$, the gate threshold $\tau$, the gate sharpness $\kappa$, and the stable-memory prior weight $w_s^0$. Within the explored neighbourhoods, all eleven cells stay within $0.94$ points of each other and within $0.58$ points of the default. NestedKV does not require per-task tuning of the surprise-router schedule.

\section{Potential Risks}

NestedKV is an inference-time efficiency method and does not train a new language model, introduce new datasets, or add new generation capabilities. Its main risk is therefore indirect: by reducing the memory cost of long-context inference, it may lower the cost of deploying existing LLMs in applications that process sensitive, copyrighted, or otherwise high-risk long documents. Any downstream deployment should inherit the safety, privacy, and data-governance controls required for the underlying model and application domain.

A second risk is reliability. KV cache compression changes the effective context available to the model, and incorrect eviction may silently remove evidence needed for faithful generation. This is particularly important in high-stakes settings such as legal, medical, financial, or security-sensitive document analysis. We therefore view NestedKV as an efficiency technique that should be validated on the target task and compression ratio before deployment, rather than as a guarantee-preserving replacement for full-cache inference.

\section{License Information}

All experiments use publicly released research artifacts, and we do not redistribute model weights or benchmark data. The Qwen3 checkpoints used in our experiments are released under Apache License 2.0. The Llama-3.2-1B-Instruct and Llama-3.2-3B-Instruct checkpoints are governed by the Llama 3.2 Community License. The \texttt{kvpress} library used for evaluation is released under Apache License 2.0.

For benchmarks, RULER's public generation code is released under Apache License 2.0. LongBench and LongBench-E are associated with the LongBench public repository under the MIT License. LooGLE's public HuggingFace dataset card lists CC-BY-SA-4.0. InfiniteBench is released under the MIT License, and MMLU-Pro is released under the MIT License. Some HuggingFace mirrors used by our runner do not specify independent license metadata; in those cases we follow the upstream benchmark licenses where available.

\end{document}